\documentclass[a4paper, 10pt, conference]{ieeeconf}      
\usepackage{FG2025}
\usepackage{tabularray}
\usepackage{graphicx}
\usepackage{booktabs}
\usepackage{amssymb}
\usepackage{multirow}
\usepackage[table]{xcolor}
\usepackage[colorlinks=true, linkcolor=red, citecolor=green, urlcolor=purple]{hyperref}
\FGfinalcopy 

\usepackage{fancyhdr}

\IEEEoverridecommandlockouts                              
\def\FGPaperID{188} 

\title{\LARGE \bf
Hands-On: Segmenting Individual Signs from Continuous Sequences
}

\author{\parbox{16cm}{\centering
   {\large Low Jian He, Harry Walsh, Ozge Mercanoglu Sincan, Richard Bowden}\\
   {\normalsize
   CVSSP, University of Surrey, Guildford, United Kingdom\\
   \texttt{\{jianhe.low, harry.walsh, o.mercanoglusincan, r.bowden\}@surrey.ac.uk}
   }}
}

\begin{document}

\ifFGfinal
\thispagestyle{empty}
\pagestyle{empty}
\else
\author{Anonymous FG2025 submission\\ Paper ID \FGPaperID \\}
\pagestyle{plain}
\fi
\maketitle

\thispagestyle{fancy}
\renewcommand{\headrulewidth}{0pt}
\fancyhf{}
\fancyhead[C]{2025 19th International Conference on Automatic Face and Gesture Recognition (FG)}


\begin{abstract}

This work tackles the challenge of continuous sign language segmentation, a key task with huge implications for sign language translation and data annotation. We propose a transformer-based architecture that models the temporal dynamics of signing and frames segmentation as a sequence labeling problem using the Begin-In-Out (BIO) tagging scheme. Our method leverages the HaMeR hand features, and is complemented with 3D Angles. Extensive experiments show that our model achieves state-of-the-art results on the DGS Corpus, while our features surpass prior benchmarks on BSLCorpus. We provide our code implementation at \url{https://github.com/JianHe0628/Hands-On}.

\end{abstract}

\section{INTRODUCTION}

Sign language is an expressive visual language that conveys meaning through hand gestures, facial expressions, and body movements \cite{signintro}. As the primary means of communication for Deaf communities, sign languages reflect both a deep cultural and linguistic identity. 

Advancements in sign language translation (SLT) systems offer a promising way to bridge communication gaps as they are designed to translate sign language into text and vice versa. However, the multi-channel nature of sign language, involving intricate hand movements, body positions, and facial expressions, poses significant challenges. This is further exacerbated by the scarcity of annotated data, particularly at the frame level, as annotating sign language videos is a labor-intensive, expertise-driven, and costly process \cite{9413817}. This limits the availability of large-scale datasets and greatly hinders the generalization of SLT systems.

Sign segmentation, the task of identifying the temporal boundaries of signs in continuous videos, is thus critical to tackle these challenges. A robust segmentation model could automate much of the gloss\footnote{A gloss is a written representation of a sign in sign language, typically using a word or phrase from the local spoken language.} annotation process, allowing linguists to refine the predicted segments instead of manually labeling them, significantly reducing dataset annotation costs. Additionally, accurate sign segmentation could enhance the performance of SLT systems. While gloss-free models have gained traction due to limited annotated data \cite{ wong2024sign2gpt, Zhou2023GlossfreeSL}, gloss-based models still consistently outperform them \cite{10.5555/3600270.3601510, zhang2023sltunetsimpleunifiedmodel}. Hence, a strong segmentation model could revitalize gloss-based approaches, unlocking new opportunities in SLT.

Prior linguistic studies have shown that sign boundaries can be identified even by non-signers, through cues such as pauses and hand lowering motions \cite{seesignboundary}. Furthermore, it was also found that over 60\% of signs in the SignBank dataset \cite{SignBankPuddle} consisted of a single hand shape \cite{moryossef2023linguistically}, highlighting the importance of hand shape features in boundary detection.

\begin{figure}
    \centering
    \includegraphics[width=1\linewidth]{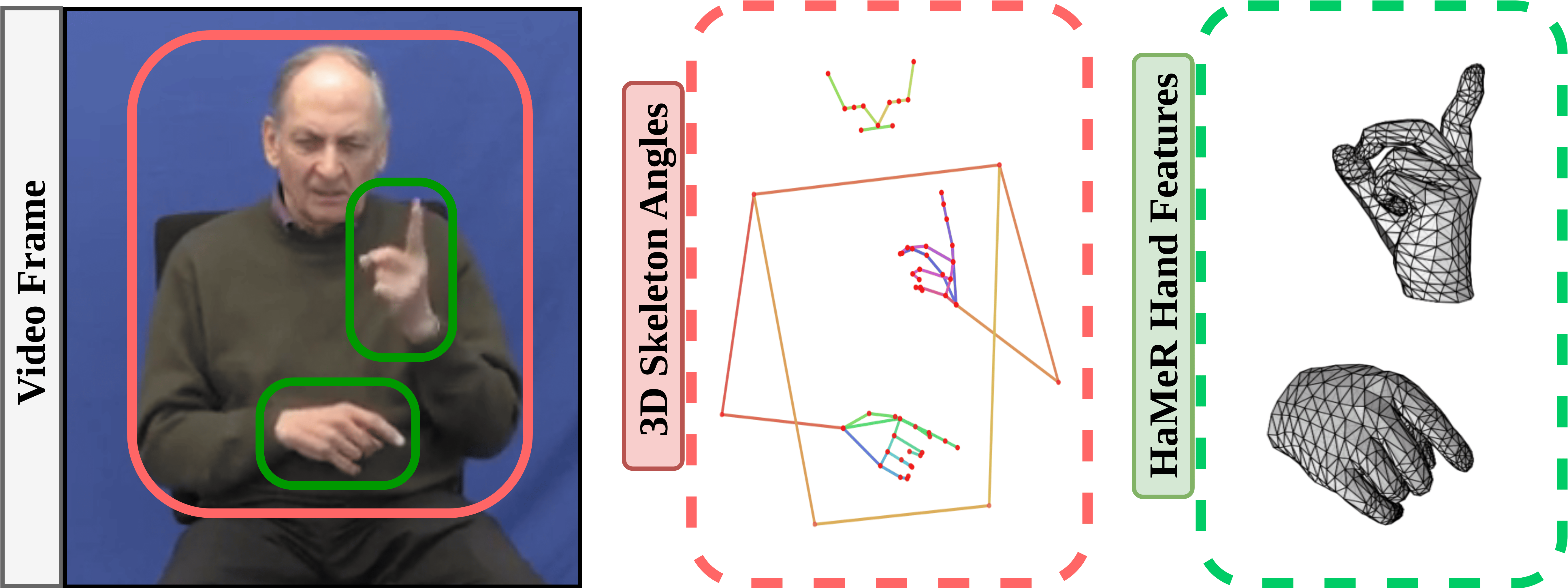}
    \caption{\textbf{Illustration of Feature Representations:} The figure showcases a sample frame from a sign language video, accompanied by its corresponding 3D skeleton-based angular pose representation, and HaMeR-generated mesh visualizations for the left and right hands.}
    \label{fig:features}
\end{figure}

Building on these insights, we propose a novel sign segmentation model that integrates robust linguistic and visual cues. Our approach leverages the state-of-the-art Hand Mesh Recovery (HaMeR) model \cite{pavlakos2024reconstructing} to extract fine-grained hand shape features and incorporates 3D skeleton angles \cite{10193629} to capture body-hand dynamics. We provide examples of these features on a single video frame in Fig. \ref{fig:features}. These features are combined into a unified multi-modal framework, which leverages a transformer-based architecture to model temporal dependencies, enabling precise identification of pauses and transitions within sign language sequences.

In summary, our contributions are as follows:
\textit{(i)} We are the first to leverage HaMeR features for sign segmentation and demonstrate their effectiveness when combined with body pose features.
\textit{(ii)} We provide extensive experiments analyzing the impact of our design choices on performance.
\textit{(iii)} We achieve state-of-the-art segmentation results on the DGS Corpus.
\textit{(iv)} We demonstrate our feature's versatility by showing they surpass prior benchmarks when integrated into alternative sign segmentation frameworks on the BSLCorpus.

\section{RELATED WORK}

Sign segmentation is the task of identifying the temporal boundaries of signs—where a sign begins and ends—within continuous sign language videos. Its difficulty lies in the ambiguity of defining precise boundaries and the labor-intensive nature of annotating frame-level data, resulting in limited datasets. Additionally, unlike spoken language, sign languages are further complicated by the simultaneous use of hand configuration, movement, location, and facial expressions \cite{Zeshan2002TowardsAN}. Despite these difficulties, significant progress has been made in sign segmentation and related tasks such as sign sequence alignment and sign localization.

Early works in sign segmentation addressed the lack of boundary annotations by aligning sign sequences to weak labels, such as TV subtitles \cite{5206523,5206647} or speech recognition sequences \cite{5457527}. These approaches relied on traditional algorithms, such as distance metrics \cite{5206523} and mining techniques \cite{5206647}. The utilization of Hidden Markov Models (HMM) in \cite{5457527} marked a significant step forward, as it formally defined the task as sign segmentation, and laid the groundwork for future research. 
Following this, more advanced machine learning techniques were then explored. For instance, a random forest classifier \cite{Farag2019LearningMD} was used to segment sign boundaries in a small-scale Japanese sign language dataset \cite{brock-nakadai-2018-deep}.

Subsequent advancements then leveraged deep learning techniques. An Inflated 3D ConvNet (I3D) \cite{8099985} with a Multi-Stage Temporal Convolutional Network (MS-TCN) \cite{8953830} was introduced in \cite{9413817} and able to achieve superior performance on the BSLCorpus \cite{Schembri2014BritishSL} and the unreleased BSL-1k dataset. Further advancements were then made in \cite{renz2021signsegmentationchangepointmodulatedpseudolabelling}, which proposed a changepoint-modulated pseudo-labeling algorithm to address source-free domain adaptation for sign segmentation. In parallel, Bull et al. \cite{9710309} enhanced I3D embeddings by integrating them with BERT \cite{devlin-etal-2019-bert} embeddings within a transformer-based architecture to perform subtitle segmentation and alignment in sign language videos. 

More recently, a Begin-In-Out (BIO) tagging scheme was introduced for sign segmentation \cite{moryossef2023linguistically}, moving beyond the traditional binary in/out frame classification. The approach highlighted the importance of prosodic cues such as facial expressions and extended pauses. Thus, it analyzed the effects of multi-modal features by combining facial landmarks, body keypoints, optical flow, and 3D hand data. This approach also established a benchmark on the MeinDGS corpus \cite{hanke-etal-2020-extending}, which has remained unchallenged. Therefore, we closely align our implementation with the methodology in \cite{moryossef2023linguistically} and evaluate our results in comparison.


\section{METHODOLOGY}

\subsection{Problem Formulation}

Prior work in sign segmentation has typically employed IO tagging or binary classification \cite{9413817}, where each frame is labeled either `in' (I) or `out' (O) of a sign, with `in' indicating that the frame is part of a sign, and `out' denoting that the frame is outside the sign. However, we instead adopt the more recently suggested formulation of sign segmentation based on a BIO tagging scheme \cite{moryossef2023linguistically}, as it yields more precise sign boundaries and has demonstrated superior performances over IO tagging across various frame rates. 

Thus, given a continuous sign language video with a sequence of frames $x = (x_1,x_2,...,x_T)$, where \textit{T} denotes the total number of frames, the goal is to produce an output sequence of $y = (y_1,y_2,...,y_T)$. Each element $y_t \in \{0,1,2\}$ corresponds to a tag indicating the phase of the sign, with 0 indicating `outside' (O), 1 for `inside' (I) and 2 for the `beginning' (B) of a sign segment.

\subsection{Features}
We hypothesize that hand shape and body position are critical features for effective sign segmentation. Thus, two carefully selected feature extraction methods were employed to capture this information:

\subsubsection{HaMeR} \label{Hamer Module}
The hand shape features are derived from the open-sourced HaMeR model \cite{pavlakos2024reconstructing}, which reconstructs 3D hand meshes. Specifically, we utilize the transformer head section of HaMeR, which was originally used to regress hand and camera parameters. In our work, we repurpose the transformer head to extract hand pose and global orientation parameters, as they were highly effective hand shape descriptors in our tests. The hand pose features are represented as a tensor of $\mathbf{H} \in \mathbb{R}^{2 \times 15 \times 3 \times 3}$, while the global orientation is parameterized as $\mathbf{G} \in \mathbb{R}^{2 \times 3 \times 3}$. For simplicity and computational efficiency, we flatten these parameters, resulting in a 1D feature vector of $\mathbf{H'} \in \mathbb{R}^{270}$ for the hand pose, and $\mathbf{G'} \in \mathbb{R}^{18}$ for the global orientation. The flattened features are then concatenated to form a feature vector of $\mathbf{F} \in \mathbb{R}^{288}$, serving as our input hand shape representation.

\subsubsection{3D Skeleton Angles} \label{angle module}
To complement hand shape information, we incorporate 3D skeleton angle features extracted using the model in \cite{10193629}. This model employs a Convolutional Neural Network (CNN) backbone, and utilizes a Forward Kinematics (FK) layer to generate a 3D human pose representation in the form of joints and angles. We adopt this model as our feature extractor due to its ability to deliver fast and accurate 3D pose estimations, while having a compact representation of $\mathbf{S} \in \mathbb{R}^{104}$. These features encode both body and hand positions, providing critical spatial context for hand movements relative to the body in 3D space.

\begin{figure*}
    \centering
    \includegraphics[width=1\linewidth]{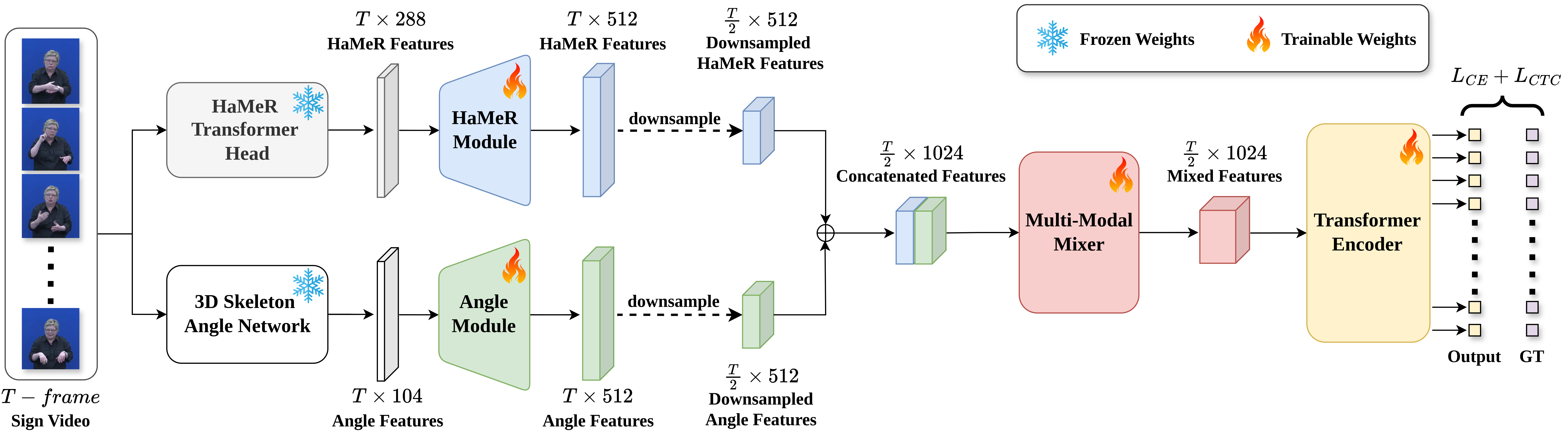}
    \caption{\textbf{Architecture Overview:} Our approach consists of three stages: \textbf{(a) Feature Extraction:} Hand shape features are extracted using the HaMeR model's transformer head, while pose features are obtained from the 3D Skeleton Angle network. Feature-specific auxiliary modules are introduced to capture the characteristics of each feature. \textbf{(b) Feature Mixing:} Extracted features are concatenated and passed through a multi-modal module to learn inter-feature relationships. \textbf{(c) Classification:} A transformer encoder processes the multi-modal features to generate (Begin-In-Out) BIO scheme predictions.}
    \label{fig:MainImplementation}
\end{figure*}

\subsection{Model Architecture} \label{Model Arch}
To achieve robust segmentation, we utilize a transformer-based architecture as the classification module due to its well-documented ability to model sequential dependencies and contextual relationships. As detailed in Sec. \ref{Hamer Module} and \ref{angle module}, hand shape and 3D skeleton features are extracted using pre-trained models; thus, they remain frozen during training. Meanwhile, to enhance feature-specific representations and facilitate multi-modal learning, we introduce two auxiliary modules and a mixer module, illustrated in Fig. \ref{fig:MainImplementation}. The auxiliary modules are feature-specific and are each a three-layer multi-layer perceptron (MLP) that up-projects HaMeR and 3D Angle features to $\mathbf{F'} \in \mathbb{R}^{512}$ and $\mathbf{S'} \in \mathbb{R}^{512}$.

To optimize computational efficiency, the features are downsampled temporally by a factor of 2. The downsampled features are concatenated along the feature dimension to form a multi-modal representation of $\mathbf{M} \in \mathbb{R}^{1024}$. The concatenated feature is then processed by the multi-modal mixer, also a three-layer MLP, to capture cross-modal interactions. The refined features are then passed to the transformer classifier to generate final per-frame predictions.

\subsection{Experimental Setup}
\subsubsection{Training}
Our model is trained for multi-class classification with target labels $y_t \in \{0,1,2\}$ using a unified loss function that combines frame-level and gloss-level objectives. For frame-level classification, we utilize multi-class cross-entropy, where $y = 2$ marks the start of a sign, $y = 1$ denotes the continuation of the sign, and $y = 0$ represents non-sign frames. Concurrently, gloss-level supervision is incorporated using the Connectionist Temporal Classification (CTC) loss \cite{10.1145/1143844.1143891}, allowing the model to leverage linguistic annotations while also providing a global constraint. 

The model is optimized using the Adam optimizer \cite{Kingma2014AdamAM} with a learning rate of $3\times10^{-4}$, and training is performed on a single RTX3090, employing the ReduceLROnPlateau scheduler with a patience of 5 epochs, gradient clipping of 0.1, and early stopping to prevent overfitting.

\subsubsection{Dataset}
\textbf{Public DGS Corpus} \cite{hanke-etal-2020-extending} is a German Sign Language dataset, recorded at 50 fps. It includes 330 sequences (average length 10 mins each) of free-flowing conversations between deaf participants on diverse topics, with precise frame-level gloss and sentence annotations.  We utilize this dataset to train our model and evaluate its performance against \cite{moryossef2023linguistically}, adhering to the MeineDGS translation protocols outlined in \cite{saunders2021signing}.

\textbf{BSLCorpus} \cite{Schembri2017BritishSL, Schembri2014BritishSL} is a British Sign Language dataset containing approximately 125 hours of video with gloss annotations and free translations. For sign segmentation, we use the subset with frame-level boundary annotations provided by \cite{9413817}, following the dataset splits and processing protocols to enable a direct comparison.

\subsubsection{Evaluation Metrics}
To ensure comparability with prior work, we evaluate our model using the same metrics proposed by \cite{moryossef2023linguistically}; specifically, the Frame-level F1 Score (F1), Intersection over Union (IoU), and Percentage of Segments (\%). The F1 Score serves as the primary criterion for both model selection and early stopping, as it provides a balanced measure of precision and recall at the frame level. The IoU metric quantifies the overlap between predicted and ground truth segments, while the \% metric complements this by assessing the ratio of predicted segments to the actual number of sign segments. Importantly, we only compute these metrics at the sign level, as phrase-level evaluation is less relevant for sign segmentation and gloss annotation. 

Additionally, we adopt the mF1B and mF1S metrics in our comparisons with \cite{9413817}. The mF1B metric assesses boundary detection by computing the F1 score across multiple integer-valued thresholds within the range of [1,4]. Meanwhile, the mF1S metric assesses segmentation quality, considering a segment correct if its IoU exceeds a specified threshold.

\section{EXPERIMENTS}

In this section, we present an ablation study for our model (Sec. \ref{Ablation Study}), our comparisons to the previous state-of-the-art (Sec. \ref{SOTA_compare}), qualitative results ( Sec. \ref{Qualitative_Results}) and our additional feature analysis on the BSLCorpus \cite{Schembri2014BritishSL} (Sec. \ref{Feature_Comparisons}).

\subsection{Ablation Study} \label{Ablation Study}

\begin{table}
\caption{Ablation Study of Features and Modules on DGS Corpus \cite{hanke-etal-2020-extending}}
\label{Ablation_Tab}
\centering
\begin{tabular}{cc|ccc|c} 
\toprule
\multicolumn{2}{c|}{Features} & \multicolumn{3}{c|}{Modules} & \multirow{2}{*}{F1}  \\  
\cline{1-5}
3D Angles & HaMeR             & 3D Angles & HaMeR & Mixer  &   \\ 
\hline\hline
\checkmark         & -                 & -         & -     & -                       & 0.8306   \\
-         & \checkmark                  & -         & -     & -                     & 0.8324   \\
\checkmark          & \checkmark                  & -         & -     & -                      & 0.8362   \\
\checkmark          & \checkmark                  & \checkmark          & \checkmark      & -                       & 0.8508   \\
\rowcolor[HTML]{ddfad3}
\checkmark          & \checkmark                  & \checkmark          & \checkmark      & \checkmark                        & \textbf{0.8566}  \\
\bottomrule
\end{tabular}
\end{table}

To understand the contribution of each design choice, we perform ablations on three key aspects: the input features, the feature-specific modules, and the mixer module. First, we evaluate the model using the Angle and HaMeR features individually, then as a fused multi-modal representation. As shown in Tab. \ref{Ablation_Tab}, our initial hypothesis that fusing HaMeR and Angle features would be the most effective is confirmed on the DGS Corpus \cite{hanke-etal-2020-extending}. While each feature performs well independently, their fusion allows the model to learn both the local hand shape information from HaMeR and the global body-hand context from the Angle features.

Additionally, the inclusion of feature-specific modules also improves performance significantly. This is likely because the modules function like adapters for the frozen feature extractors, learning representations that are tailored for sign segmentation. Finally, the mixer module's introduction also yields marginal gains, demonstrating its role in capturing interactions between modalities. It is worth noting that replacing the MLP-based mixer with a cross-attention module led to reduced performance. This performance degradation may stem from the disparity between body pose and hand shape features, causing cross-attention to introduce unnecessary complexity and hindering learning.

\begin{table} 
\centering
\caption{Comparisons to State-of-the-Art on DGS Corpus \cite{hanke-etal-2020-extending}}
\label{Sota_table}
\begin{tabular}{lccc} 
\toprule
                 & \multicolumn{3}{c}{Sign-level}  \\ 
\cline{2-4} Method & F1   & IoU & \%              \\ 
\hline\hline
Bi-LSTM + 3D Pose \cite{moryossef2023linguistically}           & 0.63 &  0.69   &   1.11              \\ 
Bi-LSTM + 3D Pose + Hand Norm \cite{moryossef2023linguistically}           & 0.59 &  0.63   &   1.13              \\ 
\rowcolor[HTML]{ddfad3}
\textbf{Ours}                            &   \textbf{0.86}   &   \textbf{0.76}  &  \textbf{0.98}               \\
\bottomrule
\end{tabular}
\end{table}

\subsection{Comparisons to State-of-the-Art} \label{SOTA_compare}

We compare our model to the current state-of-the-art approach \cite{moryossef2023linguistically} on the DGS Corpus \cite{hanke-etal-2020-extending}, which employs MediaPipe Holistic's 3D pose as input to a Bi-LSTM. As shown in Tab. \ref{Sota_table}, our model performs significantly better across all evaluation metrics. Our higher F1 score shows that we have superior frame-level sign segmentation, while the improved IoU highlights our stronger segment localizations. Additionally, our model’s score on the \% metric also approaches the ideal value of 1.00, indicating its ability to avoid excessive over and under segmentation. Furthermore, unlike \cite{moryossef2023linguistically}, who reported performance degradation when incorporating 3D hand normalizations as hand shape features, our ablation results (Tab. \ref{Ablation_Tab}) reveal that HaMeR outperforms pose-based features, showcasing its effectiveness as a robust hand shape representation for sign language tasks.

\subsection{Qualitative Results} \label{Qualitative_Results}
\begin{figure}
    \centering
    \includegraphics[width=1\linewidth]{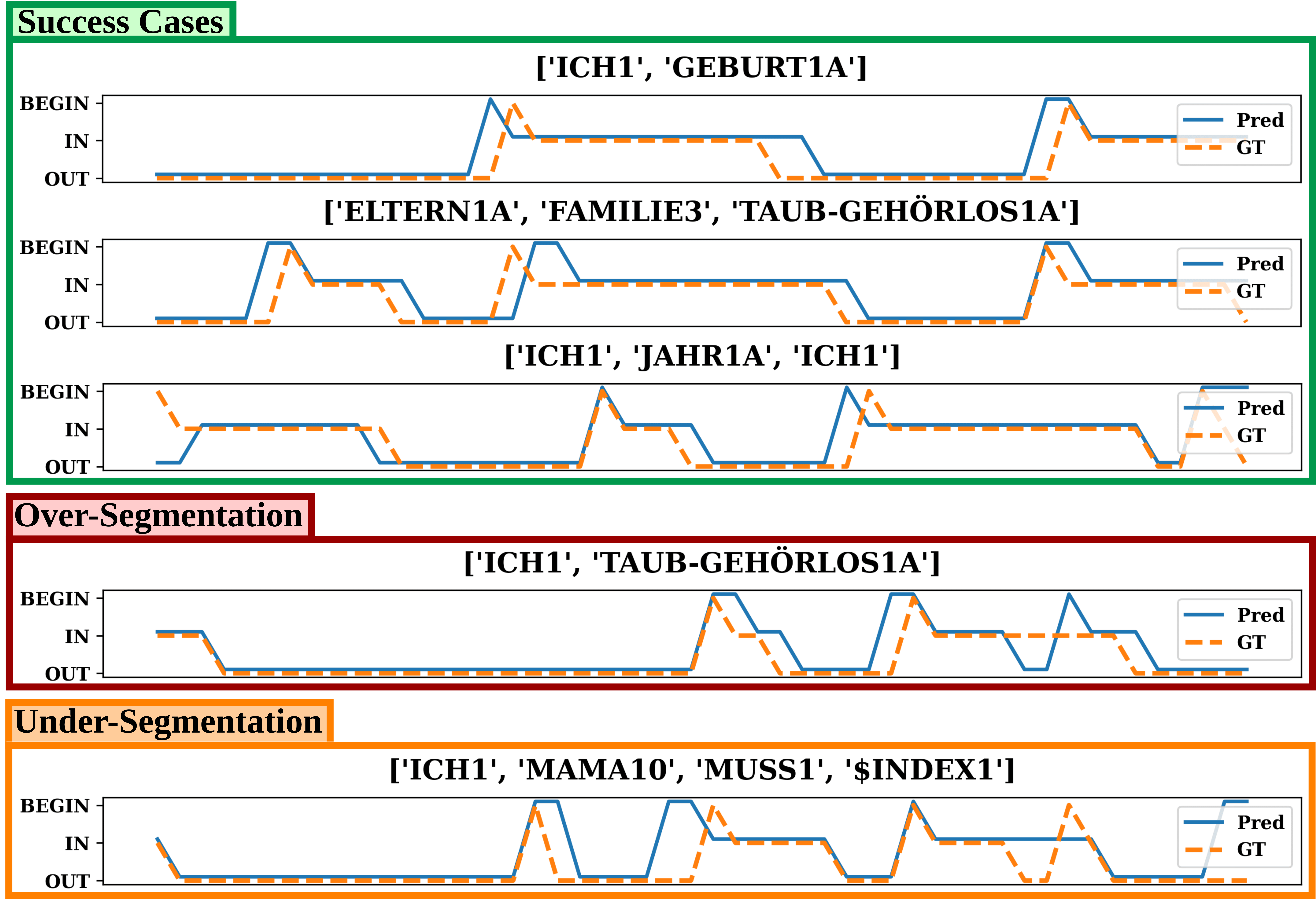}
    \caption{\textbf{Qualitative Analysis:} We provide the qualitative results of our model on the DGS Corpus \cite{hanke-etal-2020-extending}, showing the success cases (green), alongside the failure cases of over-segmentation (red) and under-segmentation (orange). Over-segmentation are instances where the model predicts a higher number of sign segments than the ground truth, while under-segmentation occurs when fewer segments are predicted than are actually present.}
    \label{fig:qual_results}
\end{figure}

In Fig. \ref{fig:qual_results}, we present qualitative results, illustrating both success and failure cases. In success cases, signs are segmented accurately, albeit with minor frame-level deviations. For instance, while the ground truth annotates a single frame as the B tag, the model occasionally extends this annotation to 2–3 frames. Slight deviations may also occur at the boundaries where segments start and end. In failure cases, we note that over-segmentation typically occurs in long signs that involve hand shape changes or pauses during signing, as these are cues the model misinterprets as transitions between signs. Meanwhile, under-segmentation occasionally arises when consecutive signs share similar hand shapes, particularly when the latter sign is considerably shorter.

\subsection{Comparisons of Features} \label{Feature_Comparisons}

\begin{table}
\centering
\caption{Feature Comparisons to State-of-the-Art on BSLCorpus \cite{Schembri2017BritishSL}}
\label{Feature_Tab}
\begin{tabular}{lcc} 
\toprule
Method & mF1B   & mF1S               \\ 
\hline\hline
MS-TCN + I3D \cite{9413817}      &   68.68   & 47.71               \\ 
MS-TCN + 3D Angles & 73.09   & 47.37 \\
MS-TCN + HaMeR & \cellcolor[HTML]{ddfad3}\textbf{76.22}   & 49.56 \\
MS-TCN + HaMeR + 3D Angles  & 75.40 & \cellcolor[HTML]{ddfad3}\textbf{50.18}   \\
\bottomrule
\end{tabular}
\end{table}

We further demonstrate the effectiveness of HaMeR and 3D Angle features for alternative sign segmentation tasks by evaluating them within the state-of-the-art framework introduced in \cite{9413817} on the BSLCorpus \cite{Schembri2017BritishSL}. The original approach leverages an MS-TCN model trained on I3D features; however, its task formulation and evaluation metrics differ significantly from ours. Thus, to ensure a direct comparison, we retrain their model using our proposed features. As shown in Tab. \ref{Feature_Tab}, HaMeR features consistently outperform I3D features, both individually and when combined with 3D Angles. Notably, HaMeR achieves superior performance despite having a feature dimensionality that is four times smaller than the I3D, demonstrating its computational efficiency and suitability for sign language tasks.

Furthermore, HaMeR and 3D Angle features exhibit greater robustness to variations in video quality and differences across human subjects. This robustness arises from their reliance on hand shapes and body poses, which are less influenced by variations in RGB pixel values compared to I3D features, making them more invariant to noise introduced by lower-quality video data and inter-subject variability.

\section{CONCLUSION}

In this work, we improved the performance of sign segmentation by proposing a transformer-based architecture trained on HaMeR and 3D skeleton angles. Extensive evaluations demonstrated that our approach outperformed state-of-the-art on the DGS Corpus. Additionally, we showed that HaMeR features generalize well to alternative sign segmentation tasks by surpassing prior benchmarks on the BSLCorpus. Future work will focus on integrating segmentation into end-to-end pipelines for sign language translation.

{ 
\section*{\small ACKNOWLEDGMENTS}
\footnotesize
This work was supported by the SNSF project ‘SMILE II’ (CRSII5 193686), the Innosuisse IICT Flagship (PFFS-21-47), EPSRC grant APP24554 (SignGPT-EP/Z535370/1) and through funding from Google.org via the AI for Global Goals scheme. This work reflects only the author’s views and the funders are not responsible for any use that may be made of the information it contains.}


\section*{ETHICAL IMPACT STATEMENT}
This research focuses on developing sign language segmentation models, which hold significant potential for advancing accessibility technologies for Deaf communities. However, it is critical to address the potential ethical concerns and risks associated with this work.

The primary risk arises from the use of publicly available datasets, namely the DGS Corpus and BSLCorpus, which include extensive video recordings of multiple signing participants. While these datasets were created with informed consent and released explicitly for research purposes, privacy concerns still remain paramount. Thus, to mitigate these risks, we strictly adhere to the licensing guidelines, ensure responsible handling of data, and avoid presenting participants in any negative or offensive manner.

Another concern is the potential for bias in the model due to the demographic imbalances in the datasets. While the DGS Corpus and BSLCorpus include diverse participants in terms of age and gender, they are predominantly European, with German and British participants, respectively. This therefore leads to under-representation of other racial and cultural groups. Unfortunately, these are the only large-scale datasets publicly available with the frame-level gloss annotations necessary for sign segmentation. However, this concern is addressed by our model's design, as it has been developed to be invariant to demographic biases. Specifically, it operates on features such as hand shape and body pose skeletons, which are independent of skin tone or facial characteristics, ensuring robustness across diverse populations.

This research brings huge societal benefits, including improved tools for sign language annotation and translation, which directly contribute to the Deaf community. Furthermore, as the datasets involved data contributed by Deaf individuals, this work directly builds upon resources provided by the community itself, ensuring relevance and inclusivity.

{\small
\bibliographystyle{IEEEtran}
\bibliography{egbib}
}

\end{document}